\documentclass[conference]{IEEEtran}
\IEEEoverridecommandlockouts

\usepackage{fancyhdr}
\usepackage{footnote}
\usepackage[hidelinks]{hyperref}
\usepackage{todonotes}
\usepackage{listings}
\usepackage{xcolor}

\usepackage{graphicx}
\usepackage{float}

\fancypagestyle{pageStyle}{%
    \fancyhf{}
    \fancyhead[C]{}

}


\ifCLASSINFOpdf
\usepackage{graphicx}
\else
\fi
\usepackage{url}

\begin{document}

%
\title{Quantifying Transparency of Machine Learning Systems through Analysis of Contributions}
\renewcommand\IEEEkeywordsname{Keywords}

\author{\IEEEauthorblockN{Iain Barclay, Alun Preece, Ian Taylor}
\IEEEauthorblockA{Crime and Security Research Institute,\\Cardiff University,\\Cardiff, UK\\
Email: BarclayIS@cardiff.ac.uk}
\and
\IEEEauthorblockN{Dinesh Verma}
\IEEEauthorblockA{IBM TJ Watson Research Center,\\1110 Kitchawan Road,\\Yorktown Heights,\\NY 10598, USA}
}


%



\maketitle
\thispagestyle{pageStyle}
\pagestyle{fancy}
\renewcommand{\headrulewidth}{0pt} 

\begin{abstract}
Increased adoption and deployment of machine learning (ML) models into business, healthcare and other organisational processes, will result in a growing disconnect between the engineers and researchers who developed the models and the model's users and other stakeholders, such as regulators or auditors. This disconnect is inevitable, as models begin to be used over a number of years or are shared among third parties through user communities or via commercial marketplaces, and it will become increasingly difficult for users to maintain ongoing insight into the suitability of the parties who created the model, or the data that was used to train it. This could become problematic, particularly where regulations change and once-acceptable standards become outdated, or where data sources are discredited, perhaps judged to be biased or corrupted, either deliberately or unwittingly. In this paper we present a method for arriving at a quantifiable metric capable of ranking the transparency of the process pipelines used to generate ML models and other data assets, such that users, auditors and other stakeholders can gain confidence that they will be able to validate and trust the data sources and human contributors in the systems that they rely on for their business operations. The methodology for calculating the transparency metric, and the type of criteria that could be used to make judgements on the visibility of contributions to systems are explained and illustrated through an example scenario.
\end{abstract}




%
\IEEEpeerreviewmaketitle

\section{Introduction}

Data from multiple sources are often aggregated and curated by human operators, and then used as the raw material for developing artificial intelligence (AI) systems. Such data-rich systems can be identified as Collective Intelligence (CI) systems, or data ecosystems, and their outputs can be new data assets or heavily data-influenced assets, including machine learning (ML) models, which are used to improve or automate decision making, either within the same organisation which created the assets or shared with partner or third party organisations.

In considering the development pipelines for ML models we can identify the contributing assets, which will typically include data sets used for training and validation, and human expertise which is used both in the preparation and curation of training data, and in the development and calibration of the resultant model. In previous work\cite{barclay2018defining},\cite{barclay2019conceptual} we have considered the benefits of applying established techniques from industry and agri-food to provide transparency and traceability on contributions to data products created through the aggregation of multiple machine and human input sources, including supply chain modelling (SCM) and the maintenance of a Bill of Materials (BoM) document to clearly identify the contributors to the output products of data ecosystems and machine learning pipelines.

Providing transparency and traceability of assets through the data supply chain or production pipeline is an important contributor to delivering accountability, which is necessary to achieve and retain confidence and trust, such that organisations using AI and data systems are able to demonstrate the provenance and authenticity of the data and knowledge they use to make decisions\cite{diakopoulos2016accountability},\cite{doshi2017accountability},\cite{weitzner2008information}. Without appropriate insight and assurance on the identity and expertise of human contributors and source or training data, an organisation could be unwittingly subject to malicious actions, including Sybil attacks\cite{wang2016defending}, data poisoning attacks\cite{miao2018towards}, and model poisoning attacks\cite{gu2017badnets}. Further, as ML models mature and are used in live production environments, it is plausible that qualifications and ethical or legal standards which were appropriate at the time the data asset or model was developed are no longer adequate by the standards in place at the time a model is used or audited, which could be many years later. The lack of transparency on the contributions to ML models and data assets is exacerbated as the distance between the developers and the users of the model increases, as is the case when models are sourced from third parties, via commercial or community marketplace platforms\cite{zhao2018packaging} and so-called model zoos\cite{jia2015caffe}.

Finding a means to quantify the overall visibility of the supply chain of data asset production pipelines is important, as it is felt that a system offering good levels of visibility on its internal workings is more likely to be considered a system affording good transparency\cite{bertino2019redefining} and supportive of achieving accountability, whereas a system with poor visibility on its contributions is likely to offer poor transparency to its end users. Systems with good transparency and accountability are likely to give better assurance on their quality and trustfulness further into the future, and thus provide a better return on investment to their developers and users. A well-regarded metric which enables systems to be rated by the transparency of their constituent components provides a mechanism for users and user communities to establish schemes to compare the transparency of models, and use this as a benchmark for building and supporting their own confidence in adopting and using a model developed by third parties, or for fostering the development of standards for good quality documentation for internal projects. Providing transparency on the contributions made to an ML model should be considered to be complementary to efforts to provide explainability\cite{preece2018stakeholders} on the outputs of AI systems, as transparency provides a means to gain assurance on the origins and builders of the system, augmenting the understanding of the system's behaviour that explainability aims to provide. 

In order to develop a quantifiable metric for the transparency of an ML production pipeline, and provide model users with a means to compare the transparency of different models, we look to the literature on supply chain visibility, particularly the work of Caridi et al\cite{caridi2010measuring},\cite{caridi2013measuring} who have developed a methodology for inferring a metric for the visibility of a manufacturing supply chain from the point of view of a focus organisation's position in the supply chain. By assigning the focus organisation viewpoint to the end user or auditor of an ML model we can assess the suitability of using Caridi's method as a basis for developing a method for ranking ML production pipelines in terms of contribution visibility, and are able to present an adaptation of Caridi's method which can be used to provide a quantifiable metric to rate the transparency of data-rich systems. 

The remainder of this paper proceeds as follows: Section \ref{sec:visibility} considers the role of supply chain visibility and its contribution towards the goals of delivering transparency and accountability, Section \ref{sec:method} provides a description of the supply chain visibility model developed by Caridi et al, and presents our contribution by way of a proposal for modifying the method so that it can be applied to ML production pipelines. Section \ref{sec:applications} details results achieved by applying the modified method to an example ML pipeline scenario when used internally and when shared with another organisations, and Section's \ref{sec:further} and \ref{sec:conclusion} respectively discuss motivations for further work and present conclusions.

\section{Visibility, Transparency and Accountability} 
\label{sec:visibility}

The study of visibility in manufacturing supply chains is an established field, which Parry et al.\cite{parry2016visibility} summarise, and identify three constructs for characterising visibility, namely ``the exchange or sharing of information", ``the properties of information exchanged" and ``the usefulness of information exchange or a capability to act on information exchange".

A discussion of visibility on published digital assets is taken up by McConaghy et al.\cite{mcconaghy2017visibility}, who make the case that the uni-directional hyper-linked nature of the world wide web leaves a lack of opportunity for dialogue between the publisher and consumer of digital assets, such that the consumer is party only to the information made available by the assert owner at publication time. If the publisher only shares a limited amount of information about the asset, then information asymmetry occurs almost by default, with the consumer of the information unaware of unreported information, such as any usage rights associated with the asset. McConaghy et al. assert that ``information availability helps both initiate and inform action, thus impacting upon an individual’s decision making process".

In seeking to provide a metric for levels of transparency on the ML production supply chain for a particular model, the emphasis is put on providing users or auditors of the output products of the systems with the capability to act upon information about the contributions to the supply chain, with the metric for visibility providing a means of determining the extent to which relevant and useful information is available to make decisions and judgements about the suitability of the system. As such, it can be argued that a production pipeline with a high-value visibility metric, relative to the scale, will provide a good level of useful information about its data sources, contributions and processes - the ``dimensions of data transparency" proposed by Bertino, et al.\cite{bertino2019redefining} - and a system with a low value visibility metric will provide a minimal amount of information, or information of low quality.

\section{Quantifying Supply Chain Visibility}
\label{sec:method}
Caridi et al. have proposed\cite{caridi2010measuring} and later refined\cite{caridi2013measuring} a method for providing a quantifiable measure for the overall visibility of a supply chain from the point of view of a focus organisation, wherein supply chain managers make semi-quantitative judgements on information available at each node in their network according to three scales - the quantity of exchanged information, the quality or accuracy of the information, and the freshness of the information. Information in the following categories are considered: transactions and events, status information, master data and operational plans, with judgements made for each of the four information categories for each node in the supply chain. Scores are awarded from 1 (lowest) to 4 (highest), when evaluated against a qualitative scale for each information category.

Caridi produces twelve judgements for each node, which are numbers from 1 to 4, according to a scale provided for each information category and each measure. Judgements assigned for information freshness and accuracy are combined to give an information quality index for each node, which is then combined with the quantity judgement to give an overall visibility rating for the node. Caridi's model then weights each node based on its closeness to the focus organisation, and its overall impact on the system, and the weighted nodes are combined to give an overall rating for the visibility of the supply chain as a number in the range from 1 for systems with the least visibility to 4 for the highest.

Caridi's method has previously been used outside its intended domain by Vlietland and van Vliet\cite{vlietland2014improving}, who adapted the model to quantify visibility of performance requirements for incident handling in IT departments. Vlietland and van Vliet used Caridi's dimensions of accuracy and freshness of information, but did not use quantity. Accuracy was used as a measure of the required and delivered performance for each node, and freshness as the timeliness of the information.

\begin{savenotes}
\vspace{5 mm} 
\begin{table*}[t]
\centering
\begin{tabular}{|l|l|l|l|}
\hline
\textbf{Score} & \textbf{Quantity} & \textbf{Freshness} & \textbf{Accuracy} \\ \hline
1              & Sparse or insufficient information          & Never updated                &  Demonstrably inaccurate        \\ \hline
2             &      Some information missing      & Out-of-date         &  Believed to be inaccurate          \\ \hline
3              & Sufficient to gain confidence              & Updated when changed            & Believed to be accurate                \\ \hline
4              & Sufficient to validate    & Real-time validation  & Evidenced and verifiable  \\ \hline
\end{tabular}
\vspace{1.5 mm} 
\caption[caption]{A scale to judge documentation on each contribution to  ML models or data assets}
\label{table:qualitativeranks}
\end{table*}
\end{savenotes}

To determine a visibility ranking for a data supply chain for a machine learning model or other produced data asset, it is proposed to use the Master Data category and adopt the same scales used in Caridi's model, but define them in a context suited to a data domain\cite{cappiello2010information}. Initially a qualitative description will be written (Table \ref{table:qualitativeranks}) to allow the scoring from 1 to 4 depending on an assessment made for each node based on interpretation and judgements made on the information in the documentation supplied with the model or data asset. When assigning scores for each of the scales, determination of the ratings should be made in regards to data sources, data sets and human participants contributing to the generation of the model or data asset, taking into consideration all information that is shared with the user or auditor of the asset. Future research will attempt to identify a set of objective measures that will be capable of being mechanised to replaces the subjective elements of the ranking. A Bill of Materials document supplied or made available with the model, as proposed\cite{barclay2019conceptual}, would be a suitable vehicle for making such information about the contributors available to model users, as it facilitates both the identification of significant contributions to the system (the nodes) as well as providing a means to identify supporting information and artifacts. Other proposals for documentation of ML systems, such as the Model Card proposal from Mitchell et al.\cite{mitchell2019model} or a Supplier's Declaration of Conformity as suggested by Hind, et al\cite{hind2018increasing}, could also be used to provide information from which the visibility judgements could be made.

\vspace{5 mm} 
\begin{table}[th]
\centering
\begin{tabular}{|l|l|}
\hline
\textbf{Judgement Criteria} & \textbf{Judgement} \\ \hline
Quantity              & $$j\textsubscript{q}$$          \\ \hline
Accuracy              & j\textsubscript{a}       \\ \hline
Freshness             & j\textsubscript{f}    \\ \hline
\end{tabular}
\vspace{1.5 mm} 
\caption[caption]{Notation for node ranking judgements}
\label{table:nodenotations}
\end{table}

By making judgements for each contributing node, \textit{k}, in the data supply chain against a set of defined criteria, as exemplified in table \ref{table:nodenotations}, it can be determined that the node's Visibility Quantity Index is:
$$VISQuantity\textsubscript{k} = j\textsubscript{q}\\$$

And for each node, \textit{k}, the Visibility Quality Index is:

$$VISQuality\textsubscript{k} = \sqrt{j\textsubscript{a} \times j\textsubscript{f}}$$

With the node's Visibility Index being:

$$VIS\textsubscript{k} = \sqrt{VISQuantity\textsubscript{k} \times VISQuality\textsubscript{k}}$$

In Caridi's model, each node is weighted according to its impact on the system, such that the overall visibility index \textit{VIS} of the supply chain is given:

$$VIS = \sum_{k=1}^{M} ( VIS\textsubscript{k} \times W\textsubscript{k} ) $$

In determining a transparency ranking for data supply chains, it is proposed to initially assign an equal weighting \textit{W\textsubscript{k}} to each contributing node, with future research considering the impact of assigning different weightings for individual contributors in different system configurations.

As such, the visibility index for a data asset or ML model resulting from a pipeline with \textit{M} contributing nodes can be determined as:

$$VIS = \sum_{k=1}^{M} \frac {VIS\textsubscript{k}} {M} $$

Where \textit{VIS} will be a number in the range 1 at the low end, to 4 at the high end for systems with very high levels of visibility on contributions towards the output. 

\section{Applications of the Method}
\label{sec:applications}
In order to assess the viability of using the proposed variant of Caridi's method to provide a quantified measure of the transparency of an ML pipeline, an example scenario illustrating the production of a simple machine learning model is rated against the information ranking criteria suggested in Table \ref{table:qualitativeranks}.

\begin{figure}[th]
\centering
\includegraphics[width=0.45\textwidth]{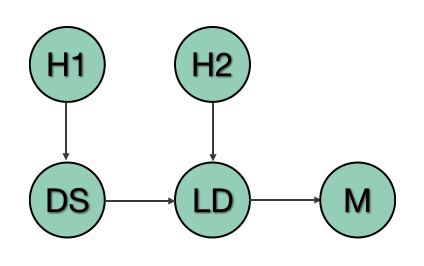} 
\caption{Production pipeline for a simple ML training process}
\label{fig:pipeline}
\end{figure}

The example pipeline (Figure~\ref{fig:pipeline}) illustrates a simple ML model training scenario, and contains a training data set (DS), which has been labelled by a single curator (H1) to produced a labelled data set (LD). An AI engineer (H2) uses the labelled data set (LD) to train and test an ML model (M), which is uploaded to a model zoo so it can be used by third parties. Note that we only need to consider the visibility of contributions from leaf nodes in the system, DS, H1 and H2 - LD is a created asset, and as such its transparency rank is determined by its contributing nodes.

To understand the impact of different levels of access to supporting information about the contributions made to the generation of the model on its transparency rating a set of scenarios are explored, these range from the point of view of the organisation which produced the model (where one would anticipate visibility would be high) to models shared with varying quality levels of supporting documentation.

\subsection{First Party Model Usage}
The first scenario considers a hypothetical model which has been generated recently, and is used in the same organisation in which it was developed, such that the users of the model can be assumed to have or to be able to gain access to complete information on the origins of the training data set and the human contributions to the system.

\vspace{5 mm} 
\begin{table}[th]
\centering
\begin{tabular}{|l|l|l|l|l|l|}
\hline
\textbf{Node} & \textbf{Quantity} & \textbf{Freshness} & \textbf{Accuracy} & \textbf{\textit{VISQuality}} & \textbf{\textit{VIS}}  \\ \hline
DS              & 4          & 4                & 4    & 4 & 4    \\ \hline
H1              & 4          & 4         & 4         & 4 & 4    \\ \hline
H2             & 4              & 4            & 4    & 4 & 4            \\ \hline
\multicolumn{5}{|l|}{Overall \textit{VIS} for model}                           & \textbf{4}                \\ \hline
\end{tabular}
\vspace{1.5 mm} 
\caption[caption]{Scores assigned to first party model usage scenario}
\label{table:firstpartyrank}
\end{table}

In such a case, it might be reasonable to apply a perfect 4 ranking to each of the judgements for each contributor to the system. Thus, from the point of view of an ML model user with good and recent knowledge of the training data, the data curator and the AI engineer responsible for training the model, it could be anticipated that the transparency rating for the model would be the maximum value of 4.

\vspace{5 mm} 
\begin{table}[th]
\centering
\begin{tabular}{|l|l|l|l|l|l|}
\hline
\textbf{Node} & \textbf{Quantity} & \textbf{Freshness} & \textbf{Accuracy} & \textbf{\textit{VISQuality}} & \textbf{\textit{VIS}}  \\ \hline
DS              & 3          & 3                & 2    & 2.45 & 2.71    \\ \hline
H1              & 3          & 3         & 3         & 3 & 3    \\ \hline
H2             & 3              & 3            & 3    & 3 & 3            \\ \hline
\multicolumn{5}{|l|}{Overall \textit{VIS} for model}                           & \textbf{2.90}                \\ \hline
\end{tabular}
\vspace{1.5 mm} 
\caption[caption]{Judgements on a first party model after a few years}
\label{table:firstpartyrankdelayed}
\end{table}

As time passes, an ML model might remain in use in an organisation but the staff responsible for the development of the model could move on to different projects or to other organisations. It may still be necessary to assess the model's transparency from time-to-time, to determine whether it is still suitable for use according to legal or ethical standards of the day, or if auditors require inspection.

In re-evaluating the information supplied with the model through documentation such as its Bill of Materials, the freshness will naturally have degraded (unless processes are in place to maintain this information and are followed adequately), and it might be that it is no longer possible to have the same confidence in the accuracy of the information supplied about the training data or the original staff involved in the curation of the data or the model training. Further, a new witness may find gaps in the information available about the contributors, such that they are unable to validate assertions made. This re-evalution of the ranking criteria might lead to new judgement scores, as in table \ref{table:firstpartyrankdelayed}, which shows a degradation in the transparency ranking of the model, as clarity and confidence in the assertions about the source data and human contributors diminishes over time, reducing the score from the optimistic 4 determined by the original creators, to a more pragmatic 2.9 rating. 

\subsection{Third Party Model Usage}
Increased sharing of ML models and generated data assets through domain-specific communities or commercial marketplaces will mean that the users of the model or the asset may have very little connection with the developers and minimal insight into how the asset was made, and in particular the qualities and qualifications of the data sources and the human operators or engineers involved. Generally this will not be problematic, however there may be instances such as inspections, audits or even legal challenges where users are required to demonstrate that their systems are suitable for use and meet the necessary regulations for their industry or domain. It is appropriate, therefore, that users of ML models and data assets are able to make an informed decision on the suitability of the systems they use by being able to form a judgement on the degree of transparency that they have on the production of the assets, such that if necessary they can trace the assets or contributors and demonstrate that they are still suitable for use. An example might be a model that was developed and published in an ML marketplace in 2019 and is still in use within an organisation in 2025, by which time it might form a critical part of an everyday business process. In performing due diligence, an auditor may seek to understand the origins of this model, such that they can gain assurance that it was produced using ethically sourced data, and that the engineers who produced it were suitably trained or qualified. The assertion being that a system with a high transparency metric will enable such checks and assurances to be efficiently made, whereas a system with low transparency may prove impossible to validate, leading to uncertainty and potentially high cost as the model is replaced.

The scenario presented here considers a model which has sparse supporting documentation, such that it is very difficult for a third party user to gain insight on the production contributors. In the worst case, it is unlikely that the user would be able to identify nodes to represent each contributor to the system, but for the sake of this example it is assumed that the model marketplace had minimal documentation requirements which ensured that such information was available to a limited extent. Reference is made to Table \ref{table:qualitativeranks} to complete the judgements in Table \ref{table:thirdpartyrank}.

\vspace{5 mm} 
\begin{table}[th]
\centering
\begin{tabular}{|l|l|l|l|l|l|}
\hline
\textbf{Node} & \textbf{Quantity} & \textbf{Freshness} & \textbf{Accuracy} & \textbf{\textit{VISQuality}} & \textbf{\textit{VIS}}  \\ \hline
DS              & 1          & 1                & 2    & 1.41 & 1.19    \\ \hline
H1              & 1          & 1         & 2         & 1.41 & 1.19    \\ \hline
H2             & 1              & 1            & 2    & 1.41 & 1.19            \\ \hline
\multicolumn{5}{|l|}{Overall \textit{VIS} for model}                           & \textbf{1.19}                \\ \hline
\end{tabular}
\vspace{1.5 mm} 
\caption[caption]{Scores assigned to minimally documented third party model usage scenario}
\label{table:thirdpartyrank}
\end{table}

Accordingly, the sparsely documented system is assigned a very low transparency score, which should serve as a warning to organisations not to allow it to become a business critical asset.

In the second instance of the scenario, consider a professionally produced and packaged ML model which comes complete with full supporting information, and perhaps a contact email address or an API to facilitate realtime queries to be made on the status of the contributing assets. Such a system might be delivered with a Model Card or a Bill of Materials document containing this information, and will provide assurance of the qualifications of the contributing staff, and information about the data sources.

\vspace{5 mm} 
\begin{table}[th]
\centering
\begin{tabular}{|l|l|l|l|l|l|}
\hline
\textbf{Node} & \textbf{Quantity} & \textbf{Freshness} & \textbf{Accuracy} & \textbf{\textit{VISQuality}} & \textbf{\textit{VIS}}  \\ \hline
DS              & 4          & 3                & 3    & 3 & 3.46    \\ \hline
H1              & 4          & 3         & 3         & 3 & 3.46    \\ \hline
H2             & 4             & 3            & 3    & 3 & 3.46            \\ \hline
\multicolumn{5}{|l|}{Overall \textit{VIS} for model}                           & \textbf{3.46}                \\ \hline
\end{tabular}
\vspace{1.5 mm} 
\caption[caption]{Judgements on a well documented third party model}
\label{table:thirdpartyrich}
\end{table}

As Table \ref{table:thirdpartyrich} demonstrates, a well-documented model from a third party source can score well for transparency, and over a number of years would perform better than a poorly documented first party system (Table  \ref{table:firstpartyrankdelayed}), as local or in-house knowledge degrades over time, which is illustrated clearly in the transparency ranking for the system. Scores could increase further with provable evidence of qualifications, but this would need to be tempered by the need to protect staff privacy and commercial secrets.

\section{Further Work}
\label{sec:further}
The implementation of the method presented applies equal weighting to each contributing node in the system, but it is arguable that some contributions have a greater impact into the final ML product or data asset than others, such as an AI Engineer configuring a system over a data curator preparing some simple and non-controversial data, and as such in some systems or use cases it may be more appropriate to weight contributions to the final transparency rank accordingly. Further, where a large team of crowd-sourced workers is deployed to curate or label the data, it may not be appropriate or necessary to have visibility on each individual contributor, and so the level of detail required will depend on the nature of the work and the domain in which the model is used. Modelling of different systems with varying types of contribution would allow exploration of the impacts of applying different weightings to contribution types on the transparency ranking of the overall system.

The criteria and scales by which transparency is determined are subjective, and would yield different results for the same systems depending on the judgement of the assessor. In order to provide a universally comparable metric, it is desirable to develop mechanisms such that objective criteria could be used to assess information quantity and quality, which could result from the use of standardised documentation requirements and formats, or in criteria and scales that could be determined mechanically. It is envisaged that the criteria and scales used will evolve within communities and organisation as users become familiar with the process of judging contributions. 

A utility of the proposed metric is to indicate to users how well they will be able to assess and have ongoing insight into systems after deployment, and after the passing of time. In order to provide the means to maintain and improve transparency rankings on systems after many years, and as they extend from the originating organisation, further work is to be conducted in finding ways to provide measurable and verifiable evidence of contributions. This could include providing mechanisms to certify the qualifications of staff, or providing systems to notify model users of any issues around the legitimacy of source data - for example, if the data set is later found to be corrupt or have contained unexpected bias which would affect the legitimacy of the model once it is in use. Blockchain platforms and schemes making use of decentralised identifiers (DIDs)\cite{reed2019} to provide self-sovereign identity (SSI)\cite{muhle2018survey} provide an interesting approach to enabling automated checks on claims and credentials to be made, whilst protecting the privacy of both the subject and the verifying party, particularly in environments where there is no direct relationship between the organisations. Further research will be conducted into this technology, with the goal of providing a mechanism to allow the highest level of transparency rating to be one that can provide irrefutable evidence of suitability of staff and data measured when against current requirements, rather than those in place at the time of model or data creation.

\section{Conclusions}
\label{sec:conclusion}
The ability to assign a quantitative ranking to the transparency of the processes and contributions which have led to the development of an ML model or data asset provides a means by which organisations can make a judgement on the suitability of a model for use in their organisation and to monitor their on-going confidence in the suitability of the model over a number of years. The strength of the subjective criteria used to assign rankings to the contributors of the asset will likely vary between organisations, but it would be helpful to see discussions on what these criteria should be, such that consensus and standard terminology can begin to emerge. Ideally, progress will be made towards determining measurable and objective criteria, such that subjective evaluations and their impact on rankings can be minimised. The use of a well understood quantitative transparency rating will become increasingly important as the distance between AI system developers and system users grows, which is inevitable as models get deployed and used both for business processes within organisations, and as they are shared and used by third parties. As such, motivation to provide and maintain accessible, up-to-date and trustable documentation and machine readable evidence of the contributions made to ML model or data asset development will become increasingly important, so as to assure users and other stakeholders that the data sets used for training the model have not been discredited, and that the staff used in data curation or engineering were appropriately trained and qualified for their tasks.

The work presented here demonstrates that guided judgements can be made on the quantity and quality of the supporting information for each contribution in a data generation pipeline and will impact the rating for the system as a whole. It is argued that the use of an adaptation of the supply chain visibility metric proposed by Caridi is of value in determining the level of transparency afforded into a data supply chain and the role of its contributors, such that systems can be evaluated and compared on the basis of their transparency. and rankings can be used as a mechanism for driving the improvement of the documentation provided with models and data assets. As further work is conducted on transparency and accountability in AI systems, it is likely that new metrics will be proposed, and it will be possible to compare and evaluate these metrics to determine which are most effective in informing stakeholders as to the levels of transparency in data ecosystems.

\section*{Acknowledgments}

This research was sponsored by the U.S. Army Research Laboratory and the UK Ministry of Defence under Agreement Number W911NF-16-3-0001. The views and conclusions contained in this document are those of the authors and should not be interpreted as representing the official policies, either expressed or implied, of the U.S. Army Research Laboratory, the U.S. Government, the UK Ministry of Defence or the UK Government. The U.S. and UK Governments are authorized to reproduce and distribute reprints for Government purposes notwithstanding any copyright notation hereon.



\bibliographystyle{IEEEtran}
\bibliography{IEEEabrv,quantifypipeline}

\end{document}